\documentclass[10pt,conference]{IEEEtran}

 \usepackage[misc]{ifsym}

\usepackage{amsmath}
\usepackage{amssymb}

\usepackage{balance}
\usepackage{booktabs}
\usepackage{epsfig}

\usepackage{stmaryrd}

\usepackage{tabularx}

\usepackage{supertabular,booktabs}
\usepackage{textcomp}

\makeatletter
\newif\if@restonecol
\makeatother

\usepackage{paralist}
\usepackage{color}
\usepackage{graphicx}
\usepackage{ifthen}
\usepackage[ruled,vlined]{algorithm2e}
\usepackage{flushend}
\usepackage{multirow}
\usepackage{xspace}
\usepackage{algorithmic}
\usepackage{amsfonts}
\usepackage{amsthm}
\usepackage{expdlist}
\usepackage{tabulary}
\usepackage{framed}
\usepackage{epstopdf}
\usepackage{colortbl}
\usepackage{amsmath}

\usepackage{caption}
\usepackage{extarrows}
\usepackage{multirow}

\usepackage{lineno}
\usepackage{longtable}

\usepackage{rotating}
\usepackage{stmaryrd}
\usepackage{subfigure}
\usepackage{tabularx}
\usepackage{graphicx}

\def\MethodName{\textsc{NSATP}\xspace}

\begin{document}

\title{A Noise-Sensitivity-Analysis-Based Test Prioritization Technique for Deep Neural Networks}
%

\author{\IEEEauthorblockN{
Long Zhang\IEEEauthorrefmark{1}\IEEEauthorrefmark{2},
Xuechao Sun\IEEEauthorrefmark{1}\IEEEauthorrefmark{2},
Yong Li\IEEEauthorrefmark{1}\IEEEauthorrefmark{2}
and
Zhenyu Zhang\IEEEauthorrefmark{2}
}
\IEEEauthorblockA{\IEEEauthorrefmark{1}University of Chinese Academy of Sciences, Beijing, China}
\IEEEauthorblockA{\IEEEauthorrefmark{2}State Key Laboratory of Computer Science, Institute of Software, Chinese Academy of Sciences, Beijing, China}

\thanks{All correspondence should be addressed to Dr. Zhenyu Zhang (Email: zhangzy@ios.ac.cn).}}

\IEEEtitleabstractindextext{
\begin{abstract}

Deep neural networks (DNNs) have been widely used in the fields such as natural language processing, computer vision and image recognition.
But several studies have been shown that deep neural networks can be easily fooled by artificial examples with some perturbations, which are widely known as adversarial examples.
Adversarial examples can be used to attack deep neural networks or to improve the robustness of deep neural networks.
A common way of generating adversarial examples is to first generate some noises and then add them into original examples.
In practice, different examples have different noise-sensitive. To generate an effective adversarial example, it may be necessary to add a lot of noise to low noise-sensitive example, which may make the adversarial example meaningless.
In this paper, we propose a noise-sensitivity-analysis-based test prioritization technique to pick out examples by their noise sensitivity.
We construct an experiment to  validate our approach on four image sets and two DNN models, which shows that examples are sensitive to noise and our method can effectively pick out examples by their noise sensitivity.

\end{abstract}
}

\maketitle

\IEEEdisplaynontitleabstractindextext
\section{Introduction}
\label{sec:intoduction}

Deep learning \cite{lecun2015deep} have been used in the fields  including molecular genetic studies \cite{xiong2015human}, brain science \cite{helmstaedter2013connectomic}, drug discovery \cite{gawehn2016deep}, medical diagnosis \cite{litjens2017survey}, mobile application \cite{han2015deep}, financial transaction \cite{deng2017deep}, and so on.
Among all the deep learning techniques, deep neural networks (DNNs) are one of the most popular techniques and have shown excellent results in  speech  recognition \cite{hinton2012deep,sainath2013deep}, image recognition \cite{schmidhuber2015deep}, natural language processing \cite{collobert2008unified}.
Many researchers have focused on developing deep neural network models \cite{he2016deep,szegedy2016rethinking,huang2017densely,krizhevsky2012imagenet}. Moreover, many open source libraries/frameworks  \cite{vedaldi2015matconvnet,jia2014caffe,abadi2016tensorflow,paszkepytorch,2016arXiv160502688short} for DNNs are available online.

In computer vision, the capabilities of deep neural network techniques  are even similar to  that of human in image recognition \cite{taigman2014deepface}.
Although there are more and more DNN models have been proposed, all of those models heavily depend on the given set of samples since DNNs are actually an effective data
representation of a given set of samples. Therefore, a trained DNN models may fail to classify a new sample correctly in practice. It is very important to know that whether a DNN
model is reliable enough to be used. There are several studies which try to expose the potential errors in a DNN model by adversarial examples. An adversarial example is an example which the given DNN models cannot classify it correctly.
%
Syegedy et al. \cite{szegedy2013intriguing} find  several interesting interesting properties of DNNs which help to launch adversarial attacks on the DNNs models for image recognition.
Later, researchers have found that they can add noise into original examples to generate adversarial examples for improving the robustness of a DNN model or attacking a DNN model.
%
%
Goodfellow et al. \cite{goodfellow2014generative} proposed a framework to estimate generative models via an adversarial process, and many variants have been proposed. They are called generative adversarial networks (GANs), which is a popular way to generate adversarial examples.
Zhang et al. \cite{zhang2018deeproad} proposed  an unsupervised framework to generate semantic-equivalent adversarial examples, which are used to test the consistency of autonomous driving systems across different scenes.
Tian et al. \cite{tian2018deeptest} proposed DeepTest, which can generate  realistic synthetic images by applying image transformations, scale, shear and rotation on original images.
Akhtari et al. \cite{akhtar2018threat} have investigated other adversarial attacks \cite{carlini2017towards,su2017one,kurakin2016adversarial,papernot2016limitations,moosavi2016deepfool,moosavi2017universal}.

To generate an effective adversarial example\footnote{In this paper, adversarial example refers to the example, which is generated by adding noise to original example, and effective adversarial example refers to the adversarial example, which can fool the deep neural network}, it needs to generate a lot of noises for adding to the
original examples before they succeed obtaining an effective adversarial example.
The number of noises to generate an adversarial example can be enormous. Moreover adversarial examples with too much noise can be easily detected by a defense model.
For example, Akhtar et al. \cite{akhtar2017defense} proposed a defense framework against the adversarial attacks generated using universal perturbations \cite{moosavi2017universal}.

In practice, we observe that different original examples can have different noise sensitivity. Examples with high noise sensitivity  are more likely to become adversarial examples by perturbations.
To increase the proportion of effective adversarial examples in all adversarial examples, high noise-sensitive example should be picked out, which are added noise to generate adversarial examples.
In this paper, we propose a noise-sensitivity-analysis-based test prioritization technique (\MethodName) to detect high noise-sensitive examples in DNN models with probability labels. First, \MethodName collects the probability vectors of original examples in deep neural network. Then, \MethodName uses the distance of the probability vectors to compute the noise-sensitive values of original examples. Finally, \MethodName ranks the original examples according to their noise-sensitive values.
We use three types of distances in this paper, namely probability difference, probability entropy and probability-variance.
We construct an experiment to validate the effectiveness of \MethodName on four image datesets and two deep neural network models. The experimental results show the example is sensitive to noise,
the higher noise sensitivity of an example is, the more \MethodName it is likely to generate an adversarial example with it, and the results based on probability difference is more effective than that based on probability entropy and probability variance.

The main contribution of our work is fourfold. First, we confirmed that the example is sensitive to noise, and adding the same noise, high noise-sensitive examples are easier to fool the deep neural network than low  noise-sensitive examples with the same noise. Second, we proposed a noise-sensitivity-analysis-based test prioritization technique, which can effectively pick out the examples by their noise sensitivity. Third, we conducted an experiment on four image datasets and two deep neural network models to validate the effectiveness of \MethodName. Fourth, probability-difference distance is more effective than probability-entropy distance and probability-variance distance.

The rest of the paper is organized as follows. Section \ref{sec:motivation} uses the observations on two images to show that the example may be sensitive to noise and motivate our work. Section \ref{sec:aprroach} presents our approach in detail, which is evaluated in Section \ref{sec:experiment}. Then  we introduce related work in Section \ref{sec:relatedwork}. Finally, we conclude our work in Section \ref{sec:conclusion}.

\section{Background and Motivation}
\label{sec:motivation}
In this section, we first present  adversarial attacks. Based on it, we discuss the observation on two image to motivate our work.

\subsection{Adversarial Attacks and Defenses}
\label{sec:motivation:attack}
 Deep neural networks are mainly driven by models and data, which are used to solve some challenging tasks. Related studies \cite{szegedy2013intriguing,goodfellow2014generative,carlini2017towards,su2017one} show that deep neural networks still have vulnerabilities, which can be fooled by  adversarial examples. There are some common ways to generate adversarial examples, such as adding noise into original examples and example transformation.

 The existence of adversarial examples has been pointed out by Szegedy et al. \cite{szegedy2013intriguing}. Given an trained model $C$, an valid input $x$  and the target $t$, some noise are added into $x$ to generate an valid input $x'$, of which the target are the same as that of $x$, but $C(x)\not= C(x')$. This example $x'$ is known as an effective adversarial example. The distance metric is used to quantify similarity between $x$ and $x'$. One main goal of related algorithms \cite{goodfellow2014generative,papernot2016limitations,carlini2017towards,kurakin2016adversarial} is to generate an adversarial example $x'$ with a small distance from $x$.

Since the discovery of adversarial examples for the deep neural networks \cite{szegedy2013intriguing}, many works focus on the robustness of neural networks against these adversarial examples and show that using adversarial examples as training can improve the robustness of neural networks. Akhtar et al. \cite{akhtar2017defense} proposed a defense framework  appending a pre-input layers against the adversarial attacks. adversarial examples with too much noise are  easily identified by defense mechanism. There are some defense mechanism that limit the number of attacks, such as Alipay's face authentications.

\begin{figure}[htbp]
\centering
\subfigure[label 4 (81\%)]{
\begin{minipage}[t]{0.25\linewidth}
\centering
\includegraphics[height=1.5cm,width=1.5cm,page=1]{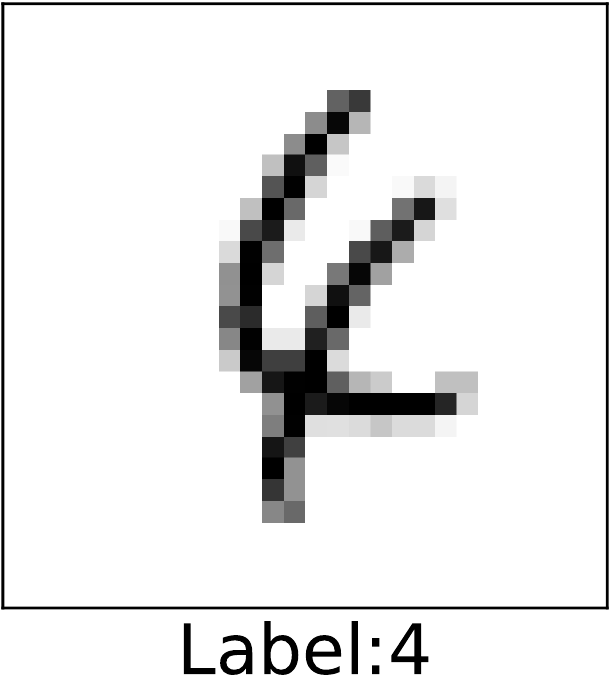}
\end{minipage}%
}%
\subfigure[label 4 (95\%)]{
\begin{minipage}[t]{0.25\linewidth}
\centering
\includegraphics[height=1.5cm,width=1.5cm,page=2]{motivation-crop.pdf}
\end{minipage}%
}
\caption{ Two Original Images}
\label{fig:original}
\end{figure}

\subsection{Observation}
\label{sec:motivation:observation}
\begin{figure}[htbp]
\centering
\subfigure[Orig. 1]{
\begin{minipage}[t]{0.16\linewidth}
\centering
\includegraphics[height=1.5cm,width=1.5cm,page=1]{motivation-crop.pdf}
\end{minipage}%
}%
\subfigure[adver. 1]{
\begin{minipage}[t]{0.16\linewidth}
\centering
\includegraphics[height=1.5cm,width=1.5cm,page=3]{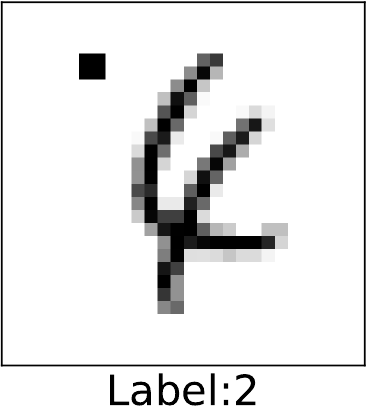}
\end{minipage}%
}%
\subfigure[adver. 2]{
\begin{minipage}[t]{0.16\linewidth}
\centering
\includegraphics[height=1.5cm,width=1.5cm,page=4]{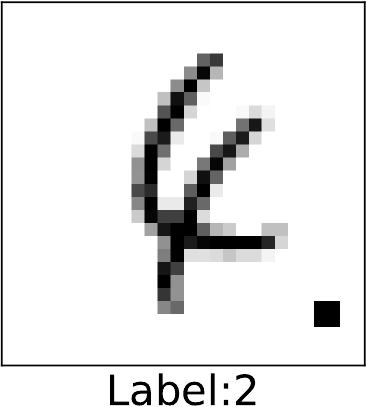}
\end{minipage}%
}%
\subfigure[adver. 3]{
\begin{minipage}[t]{0.16\linewidth}
\centering
\includegraphics[height=1.5cm,width=1.5cm,page=5]{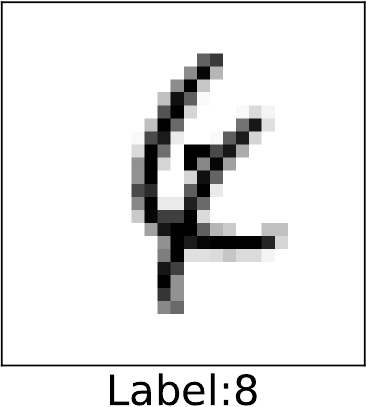}
\end{minipage}
}%
\subfigure[adver. 4]{
\begin{minipage}[t]{0.16\linewidth}

\includegraphics[height=1.5cm,width=1.5cm,page=6]{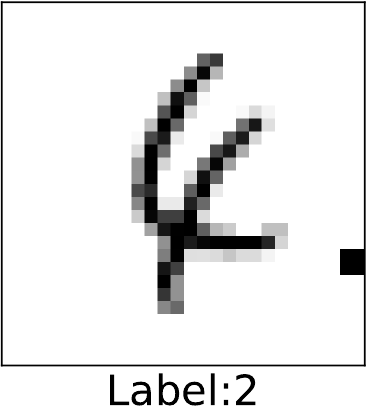}
\end{minipage}
}%
\\
\subfigure[adver. 5]{
\begin{minipage}[t]{0.16\linewidth}
\centering
\includegraphics[height=1.5cm,width=1.5cm,page=7]{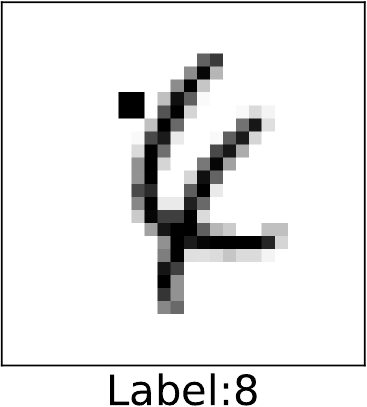}
\end{minipage}%
}%
\subfigure[adver. 6]{
\begin{minipage}[t]{0.16\linewidth}
\centering
\includegraphics[height=1.5cm,width=1.5cm,page=8]{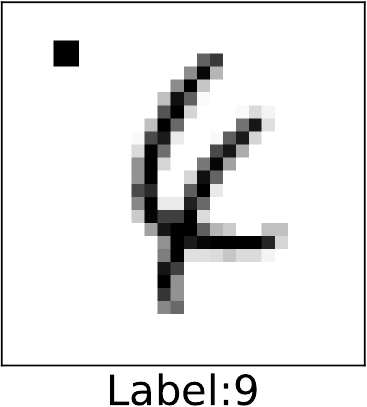}
\end{minipage}%
}%
\subfigure[adver. 7]{
\begin{minipage}[t]{0.16\linewidth}
\centering
\includegraphics[height=1.5cm,width=1.5cm,page=9]{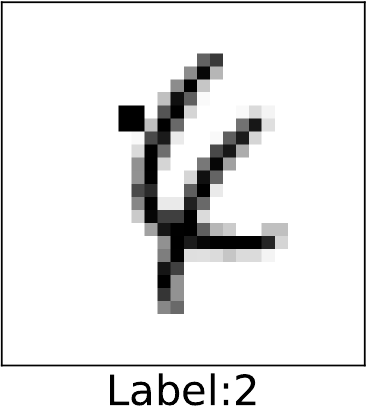}
\end{minipage}%
}%
\subfigure[adver. 8]{
\begin{minipage}[t]{0.16\linewidth}
\centering
\includegraphics[height=1.5cm,width=1.5cm,page=10]{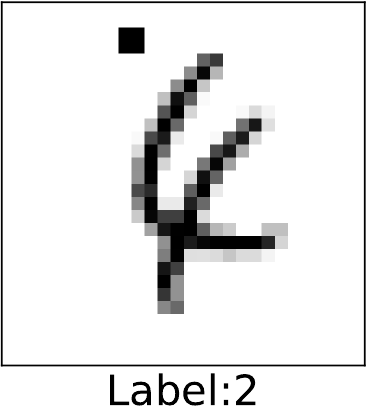}
\end{minipage}%
}%
\centering
\caption{Adversarial examples of Fig.~\ref{fig:original} (a)}
\label{fig:adv1}
\end{figure}

\begin{figure}[htbp]
\centering
\subfigure[Orig. 2]{
\begin{minipage}[t]{0.16\linewidth}
\centering
\includegraphics[height=1.5cm,width=1.5cm,page=2]{motivation-crop.pdf}
\end{minipage}%
}%
\subfigure[adver. 1]{
\begin{minipage}[t]{0.16\linewidth}
\centering
\includegraphics[height=1.5cm,width=1.5cm,page=30]{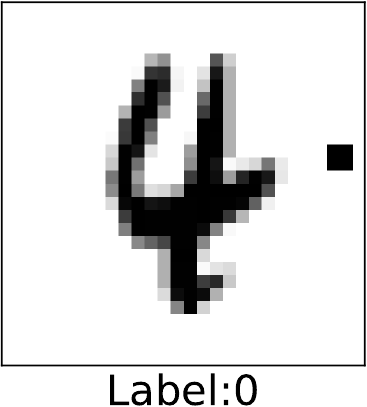}
\end{minipage}%
}%
\subfigure[adver. 2]{
\begin{minipage}[t]{0.16\linewidth}
\centering
\includegraphics[height=1.5cm,width=1.5cm,page=31]{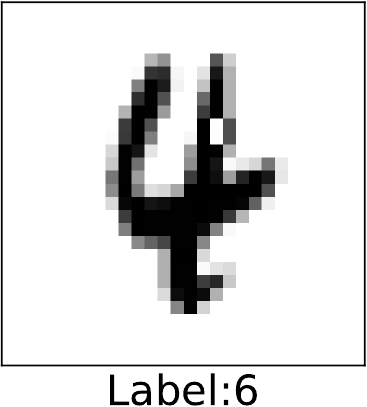}
\end{minipage}%
}%
\\
\subfigure[adver. 3]{
\begin{minipage}[t]{0.16\linewidth}
\centering
\includegraphics[height=1.5cm,width=1.5cm,page=32]{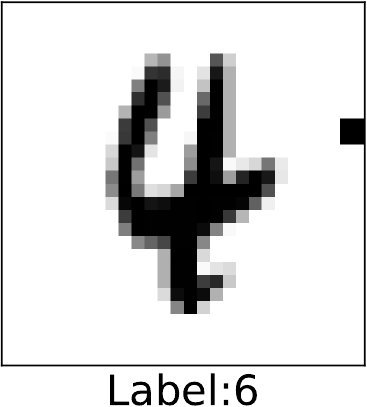}
\end{minipage}
}%
\subfigure[adver. 4]{
\begin{minipage}[t]{0.16\linewidth}

\includegraphics[height=1.5cm,width=1.5cm,page=33]{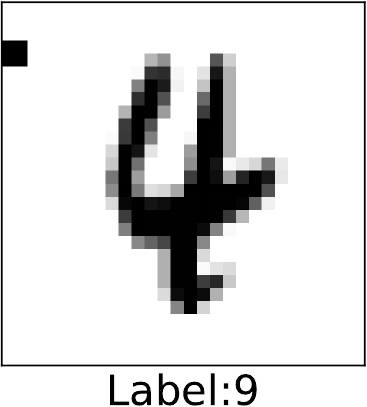}
\end{minipage}
}%
\subfigure[adver. 5]{
\begin{minipage}[t]{0.16\linewidth}
\centering
\includegraphics[height=1.5cm,width=1.5cm,page=34]{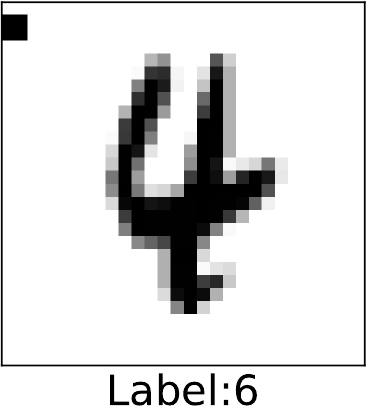}
\end{minipage}%
}%

\centering
\caption{Adversarial Examples of Fig.~\ref{fig:original} (b)}
\label{fig:adv2}
\end{figure}

\begin{figure*}[htbp]
\centering
\includegraphics[width=16cm,page=2]{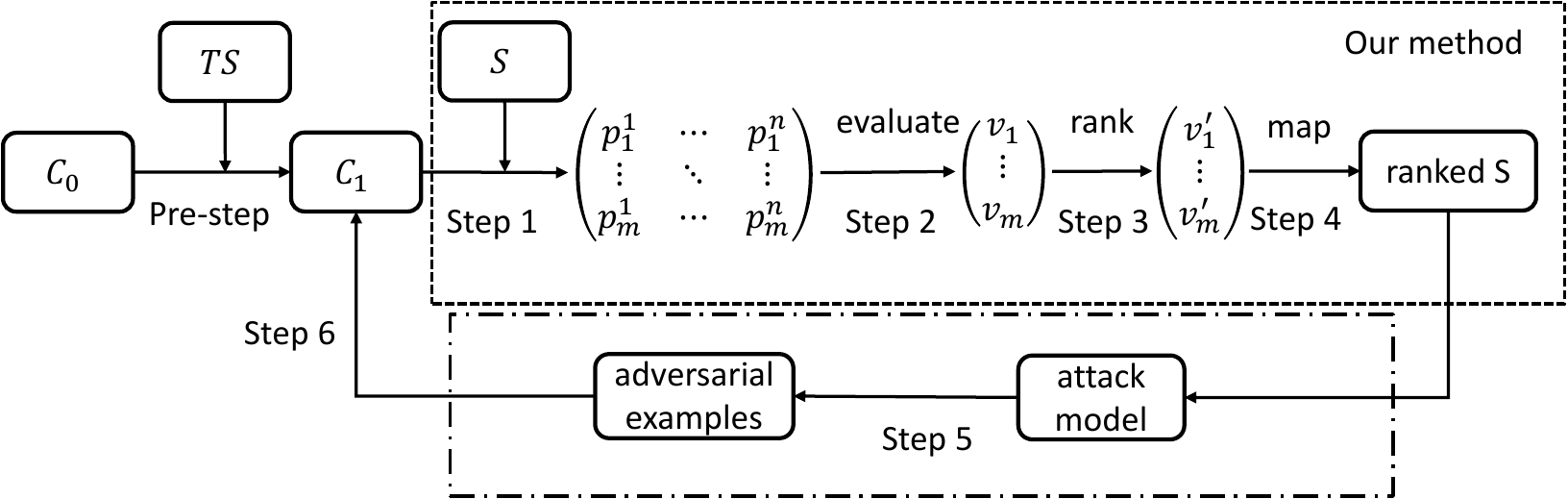}
\caption{Workflow of Our Method}
\label{fig:workflow}
\end{figure*}

Convolutional neural networks are  a special kind of multi-layer neural networks, which are trained with a version of the back-propagation algorithm. LeNet \cite{lecun1995comparison} is one convolutional network designed for handwritten and machine-printed character recognition.

We use MNIST \cite{lecun1998mnist} to train LeNet-5 model, and randomly choose two pictures showed in Fig.~\ref{fig:original}. LeNet-5 outputs that the probability vectors of the two pictures are $\langle0,0,0,0.18,0.81,0,0,0,0.01,0\rangle$ and $\langle 0,0,0,0,0.95,0.01,0,0,0.01,0.03\rangle$, respectively. We randomly choose $2\times2$ pixels and reverse the value of each pixel in each picture. We repeat it to generate 50 adversarial examples for each pictures. There are 8 and 5 effective adversarial examples with respect to the two pictures of Fig.~\ref{fig:original}, showed in Fig.~\ref{fig:adv1} and Fig.~\ref{fig:adv2}. We have the observation that the number of effective adversarial examples with respect to Fig.~\ref{fig:original} (a) is bigger than that with respect to Fig.~\ref{fig:original} (b). We repeat 10 times to observe the difference between the numbers of effective adversarial examples with respect to Fig.~\ref{fig:original} (a) and Fig.~\ref{fig:original} (b), and we find the numbers of effective adversarial examples with respect to Fig.~\ref{fig:original} (a) is always not less than that with respect to Fig.~\ref{fig:original} (b).
We further observe the probability vectors of the two pictures, and the probability values of each label of Fig.~\ref{fig:original} (b) are  more discretized than that of Fig.~\ref{fig:original} (a). For example, the probability vector variance (0.80) of Fig.~\ref{fig:original} (b) is more than that (0.59) of Fig.~\ref{fig:original} (a), and the probability difference ($0.95-0.03=0.92$)  of Fig.~\ref{fig:original} (b) between 1st probability and 2nd probability is more than that ($0.81-0.18=0.73$) of Fig.~\ref{fig:original} (a).
The observation shows that  the examples are sensitive to noise, which motivates us to consider whether the noise sensitivity is related to probability vectors, and whether probability vectors can be used to evaluate the noise sensitivity of given examples.

\section{Technique Framework}
\label{sec:aprroach}
In this section, we present our method \MethodName to compute the noise sensitivity of each examples and rank examples by their noise sensitivity. Our method is designed for choosing the original examples before generating adversarial examples based on  original examples to attack deep neural networks or improve the robustness of deep neural networks.

\subsection{Problem Settings}
A neural network can be seen as a function $F(x)=y$, where x denotes the input and y denotes the output ($x\in R$ and $y\in R$).
In this paper we focus on deep neural networks with the capabilities of n-class classification. The networks use  the softmax function to compute the output of the network, and the output vector $p$ satisfies $ p^i \ge 0$  and $\sum^{i=n}_{i=1}p^i = 1$.

Let us consider  a train set TS and a deep neural network models $C_0$, of which the outputs are the probability labels.  All examples of TS can be used to train the model $C_0$ to generate a trained a trained model $C_1$ with n probability vectors. For a valid input example $t_i \in TS$, the trained model $C_1$ can output the probability vector of $t_i$, i.e., $\langle p^1_i, p^2_i, \ldots, p^n_i\rangle$.

Suppose that there is a example set S (i.e., $\langle s_1, s_2, \ldots, s_m\rangle$), which is used to generate adversarial examples, and the method $AF$ of adding noise  to S is available. For a example $s_i \in S$, $AF$ can generate $k (k \geq 1)$ examples (i.e., $\langle s^1_i, s^2_i, \ldots, s^k_i\rangle$), which are used to attack the trained model $C_1$ or improve the robustness of the trained model $C_1$.

\subsection{Our Method \MethodName}

Since our method (named \MethodName) is to evaluate the noise sensitivity of each example in S based on their probability vectors computed by the trained model $C_1$, the probability vectors of evaluated examples should be available. Our method workflow is depicted in Fig. \ref{fig:workflow}.

Previous related works \cite{szegedy2013intriguing,goodfellow2014generative,carlini2017towards} focused on how to generate effective adversarial examples based on one given example, but example may be  sensitive to noise.  adding too much noise to low noise-sensitive example to generate effective adversarial examples  may make the adversarial example meaningless or be easily detected by the defense mechanism. \MethodName aims to assess the noise sensitivity of examples and pick out the high noise-sensitive examples before attack models generate adversarial examples.

To start such a process, we assume that a trained models $C_1$ is ready showed in pre-step of Fig. \ref{fig:workflow}, and  the examples $S$ is also ready before applying \MethodName. \MethodName four steps, as illustrated in the Fig. \ref{fig:workflow}.
\begin{enumerate}[\ \ \ $Step$ 1:]
\item Input $S$ into $C_1$ to collect the probability vectors of each $s_i \in S$;
\item Use a distance metric to evaluate the probability vectors of each $s_i$;
\item Generate a ranking list based on the results of $step$ 2;
\item Map the ranking list of $step$ 3 and $S$ to generate the ranked example list based on $S$.
\end{enumerate}
Previous related works \cite{szegedy2013intriguing,goodfellow2014generative,carlini2017towards} focused on the step 5 of Fig. \ref{fig:workflow}, which randomly selected examples from $S$ and ignored the noise sensitivity of examples in $S$. Then, they used the generated adversarial examples to attack $C_1$ or improve the robustness of $C_1$.

\begin{equation}
  \label{eq:pd}
  PD=\sum_{i=1}^{i=n-1}\frac{(p'^i -p'^{i+1})}{i}
\end{equation}
\begin{equation}
  \label{eq:pv}
  PV=\sum_{i=1}^{i=n}(p'^i -\overline{p})^2
\end{equation}
\begin{equation}
  \label{eq:pe}
  PE=-\sum_{i=1}^{i=n}p'^i\log p'^i
\end{equation}

The selection of distance metrics is crucial to our method. Based on the observation in Section \ref{sec:motivation:observation}, the distance metric is evaluated the noise sensitivity of one example based on the difference between the probability vectors of the example. Suppose we are given one probability vector $\langle p^1, p^2, \ldots, p^n\rangle$, we use probability difference (PD), probability variance (PV) and probability entropy (PE) to compute the noise sensitivity, respectively. The probability difference is the sum of the differences between different sorting levels with different weights.
We first rank each value of the  given probability vector to generate a ranked vector $\langle p'^1, p'^2, \ldots, p'^n\rangle$ by descending order, then compute the difference between two adjacent numbers of the ranked vector and  mark the weight from $\frac{1}{1}$ to  $\frac{1}{n-1}$, finally calculate the sum of all difference with their weights.  The PD calculation is given in Equation \ref{eq:pd}. The variance is used to measure the dispersion of one given data set. We use the variance metric to measure the dispersion between values in one given probability vector. The PV calculation is given in Equation \ref{eq:pv}. Entropy can be used to evaluate the uncertainty of random variables, and the probability values of the example can  be seen as random variables. In this paper we call it probability entropy. The PE calculation is given in Equation \ref{eq:pe}.

\section{Evaluation}
\label{sec:experiment}
In this section, we first describe set the experiment and train the models on which we evaluate our method. Then, we introduce some metrics to evaluate the effectiveness of our method. Finally, we present  the data analysis and the results of the experiment.

\subsection{Research Questions}
Regarding our obversactions in Section \ref{sec:motivation} and our proposal, we are desired to answer the
following questions.
\begin{enumerate}[\ \ \ $Q$ 1:]
\item Whether are examples sensitive to noise, and is there a relationship between  noise sensitivity and probability vectors?
\item If Q2 is "yes", is our method effective to analyze the noise sensitivity of examples than  the standard random manner?
\item If Q3 is "yes", which distance is better to evaluate the the noise sensitivity of examples?
\end{enumerate}

\subsection{Experimental Setup}
\begin{table*}[htbp]
  \caption{Model Architectures}
  \label{tab:model}
  \centering
  \begin{tabular}{l|cccc|cccc}
  \toprule
  & \multicolumn{4}{ c |}{ Net-1} & \multicolumn{4}{ c }{ Net-2} \\
  Layer Type & MNIST & f-MNIST & CIFAR-10 & CIFAR-100 & MNIST & f-MNIST & CIFAR-10 & CIFAR-100\\
  \midrule
  Convolution + ReLU &  $5\times 5\times 64$             &    $5\times 5\times 64$      &     $5\times 5\times 64$      &   $5\times 5\times 64$     &      $3\times 3\times 32$        &   $3\times 3\times 32$       &    $3\times 3\times 32$ &    $3\times 3\times 32$    \\
  Max Pooling        &   $3\times 3$            &    $3\times 3$      &     $3\times 3$      &    $3\times 3$    &     $2\times 2$         &    $2\times 2$      &     $2\times 2$  &   $2\times 2$   \\
   Convolution + ReLU &  $5\times 5\times 64$             &    $5\times 5\times 64$      &     $5\times 5\times 64$      &   $5\times 5\times 64$    &      $3\times 3\times 32$        &   $3\times 3\times 32$       &    $3\times 3\times 32$ &    $3\times 3\times 32$    \\
    Convolution + ReLU &  $5\times 5\times 64$             &    $5\times 5\times 64$      &     $5\times 5\times 64$      &   $5\times 5\times 64$     &      -        &   -       &    - &    -    \\
  Max Pooling       &   $3\times 3$            &    $3\times 3$      &     $3\times 3$      &    $3\times 3$    &     $2\times 2$         &    $2\times 2$      &     $2\times 2$  &   $2\times 2$   \\
  Fully Connected + ReLU&     512          &    512      &     512      &    512    &     512         &    512      &     512   &   512   \\
  Fully Connected + ReLU&      192         &    192      &     192      &    192    &      -        &   -       &   -   & -     \\
  SoftMax            &      10         &     10     &    10       &    100    &     10         &   10       &    10 & 100    \\
  \bottomrule
  \end{tabular}
\end{table*}%

\begin{table}[htbp]
  \caption{Model Architectures}
  \label{tab:para}
  \centering
  \begin{tabular}{l|cccc}
  \toprule
  Layer Type & MNIST & f-MNIST & CIFAR-10 & CIFAR-100 \\
  \midrule
  learning rate & 0.1 & 0.1 & 0.01 & 0.01\\
  Batch Size    & 512 & 512 & 512  & 512 \\
  Epochs        & 20  & 20  & 60   &  60 \\
  \bottomrule
  \end{tabular}
\end{table}%
Similar to Su et al. \cite{su2017one}, we encode the perturbation into an array which is optimized by differential evolution. In our experiment each perturbation holds  x-y coordinates and RGB value of the perturbation, and one perturbation modifies four ($2\times 2$) pixels.

We train two network models on  MNIST, fashion-MNIST (Shorted in f-MNIST),  CIFAR-10 and CIFAR-100 classification tasks, respectively. The model architecture is given in
Table \ref{tab:model} and the hyperparameters selected in Table II. In our paper, we use the all training sets of the four datasets to traning DNN models, and use the test sets of the four datasets to analyze the noise sensitivity of examples and answer the research questions.

\subsection{Performance Measurement}
In this section, we will introduce some metrics used to evaluate the effectiveness of \MethodName.

\subsubsection {Effective adversarial sample ratio (Eff.)} Evaluating the effectiveness of adversarial examples is an important way to compare the effectiveness of different adversarial attack techniques. The more  the effectiveness of adversarial examples is, the less cost it is to successfully attack DNNs.
In this paper, to measure the noise-sensitivity of examples, we compute how many adversarial samples fool the given deep neural network among given adversarial sample set. For example, given an example $s$ and a deep neural network, an attack model generate m adversarial samples, of which n samples can fool the given deep neural network, and the nosie-senstivity of $s$ is $\frac{n}{m}$.

\subsubsection{ F-measure} F-measure \cite{chen2004adaptive} is widely used to evaluate the effectiveness of different methods in adaptive testing and security testing. F-measure denotes how many test cases are executed before successfully attacking the system.  To measure performance the advantages in effectiveness of NSATP to Random, we follow \cite{carlini2017towards,chen2004adaptive} to use the F-measure metric. F-measure calculates the expected number of adversarial sample required to successfully fool the given deep neural network. In other words, a lower F-measure value means that fewer adversarial samples are used to accomplish the task. If an attack strategy yields a lower F-measure value, it is considered to be more effective. We expect a lower F-measure value for NSATP than that of Random. We thus calculate $Inc.=\frac{R-A}{R}\times 100\%$ to evaluate the effectiveness improvement, where R and A stand for the effectiveness of Random and NSATP in F-Measure, respectively.  The greater the value is, the more effective NSATP is than Random.

\subsubsection{ Effectiveness on distance metric(Eff-D)} In our framework, it can use any distance metric, which can be used to evaluate the probability vectors. In this paper, we use three different distance metrics to compute the noise sensitivity of given examples. To evaluate the effectiveness of different ranks, we can fit their curve functions and map them to linear functions, then use the absolute derivative of linear functions to evaluate the effectiveness of different ranks. First, for a given distance metric, we use our framework to obtain a tuple $\langle v, E\rangle$ for an given example, where $v$ denotes the distance value of the probability vector of the given example and $E$ denotes the noise sensitivity of the given example. Then, we use the tuple set to fit a curve function. For different distance metric on one example set, we can fit different curve with 95\% correlation coefficient. Third, we map the curve function to a linear function, and compute the derivative of each fitted linear function. Finally, we compare the absolute values of different derivatives to analyze which distance metric is better. The more the absolute value is, the better the corresponding distance metric is.

\subsection{Data Analysis and Results}

We report our experiment results in three subsections to answer the three research questions, respectively.

\subsubsection{Answer Q1}

\begin{figure}[htbp]
\centering
\subfigure[MNIST on PD]{
\begin{minipage}[t]{0.31\linewidth}
\centering
\includegraphics[width=2.9cm,page=1]{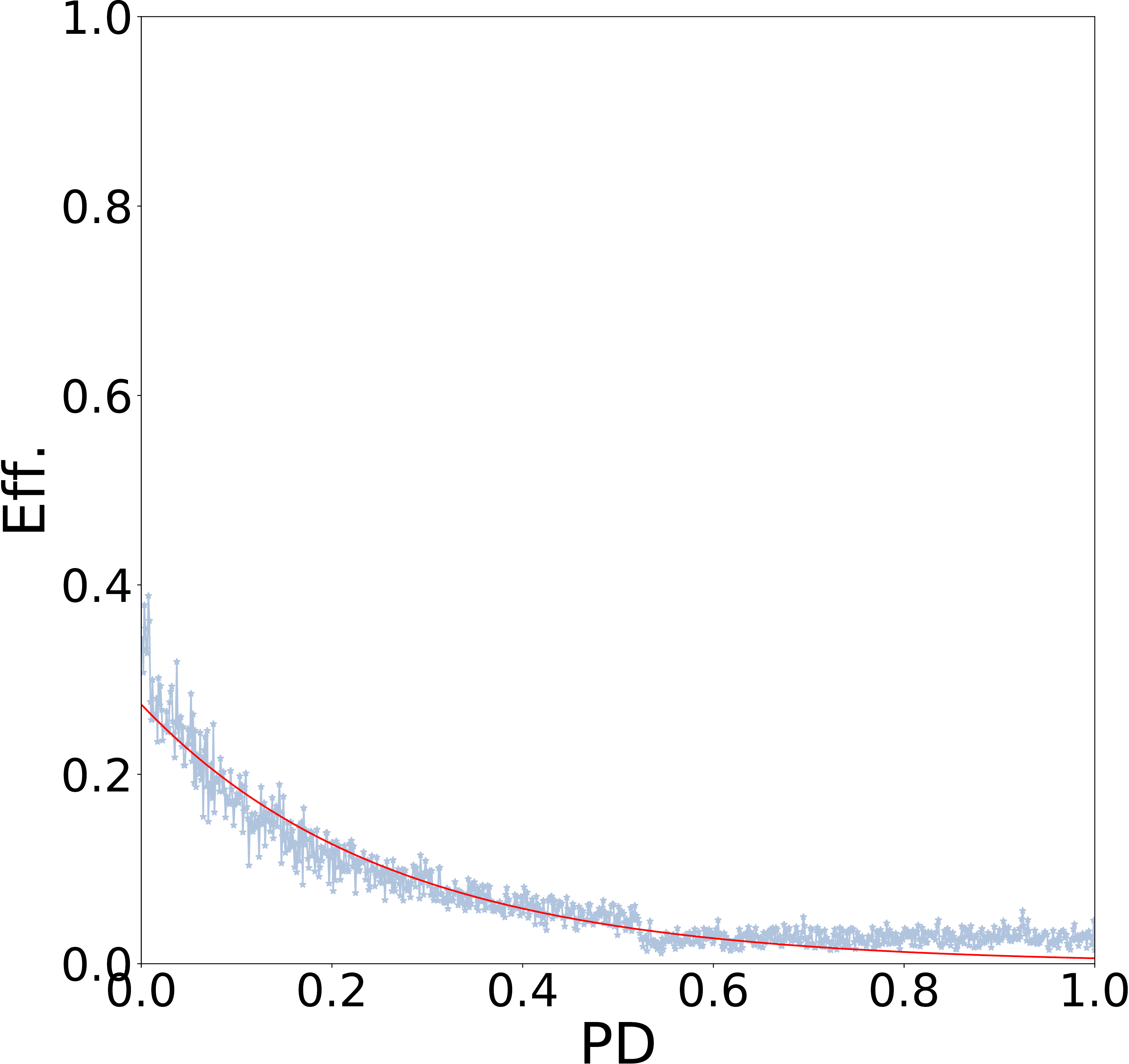}
\end{minipage}%
}%
\subfigure[f-MNIST on PD]{
\begin{minipage}[t]{0.31\linewidth}
\centering
\includegraphics[width=2.9cm,page=2]{figure_lenet.pdf}
\end{minipage}%
}
\subfigure[Cifar-10 on PD]{
\begin{minipage}[t]{0.31\linewidth}
\centering
\includegraphics[width=2.9cm,page=3]{figure_lenet.pdf}
\end{minipage}%
}
\\
\subfigure[Cifar-100 on PD]{
\begin{minipage}[t]{0.31\linewidth}
\centering
\includegraphics[width=2.9cm,page=4]{figure_lenet.pdf}
\end{minipage}%
}%
\subfigure[MNIST on PE]{
\begin{minipage}[t]{0.31\linewidth}
\centering
\includegraphics[width=2.9cm,page=5]{figure_lenet.pdf}
\end{minipage}%
}
\subfigure[f-MNIST on PE]{
\begin{minipage}[t]{0.31\linewidth}
\centering
\includegraphics[width=2.9cm,page=6]{figure_lenet.pdf}
\end{minipage}%
}
\\
\subfigure[Cifar-10 on PE]{
\begin{minipage}[t]{0.31\linewidth}
\centering
\includegraphics[width=2.9cm,page=7]{figure_lenet.pdf}
\end{minipage}%
}%
\subfigure[Cifar-100 on PE]{
\begin{minipage}[t]{0.31\linewidth}
\centering
\includegraphics[width=2.9cm,page=8]{figure_lenet.pdf}
\end{minipage}%
}
\subfigure[MNIST on PV]{
\begin{minipage}[t]{0.31\linewidth}
\centering
\includegraphics[width=2.9cm,page=9]{figure_lenet.pdf}
\end{minipage}%
}
\subfigure[f-MNIST on PV]{
\begin{minipage}[t]{0.31\linewidth}
\centering
\includegraphics[width=2.9cm,page=10]{figure_lenet.pdf}
\end{minipage}%
}
\subfigure[Cifar-10 on PV]{
\begin{minipage}[t]{0.31\linewidth}
\centering
\includegraphics[width=2.9cm,page=11]{figure_lenet.pdf}
\end{minipage}%
}
\subfigure[Cifar-100 on PV]{
\begin{minipage}[t]{0.31\linewidth}
\centering
\includegraphics[width=2.9cm,page=12]{figure_lenet.pdf}
\end{minipage}%
}
\caption{ The Results of Net-2}
\label{fig:resutlsNet2}
\end{figure}

\begin{figure}[htbp]
\centering
\subfigure[MNIST on PD]{
\begin{minipage}[t]{0.31\linewidth}
\centering
\includegraphics[width=2.9cm,page=1]{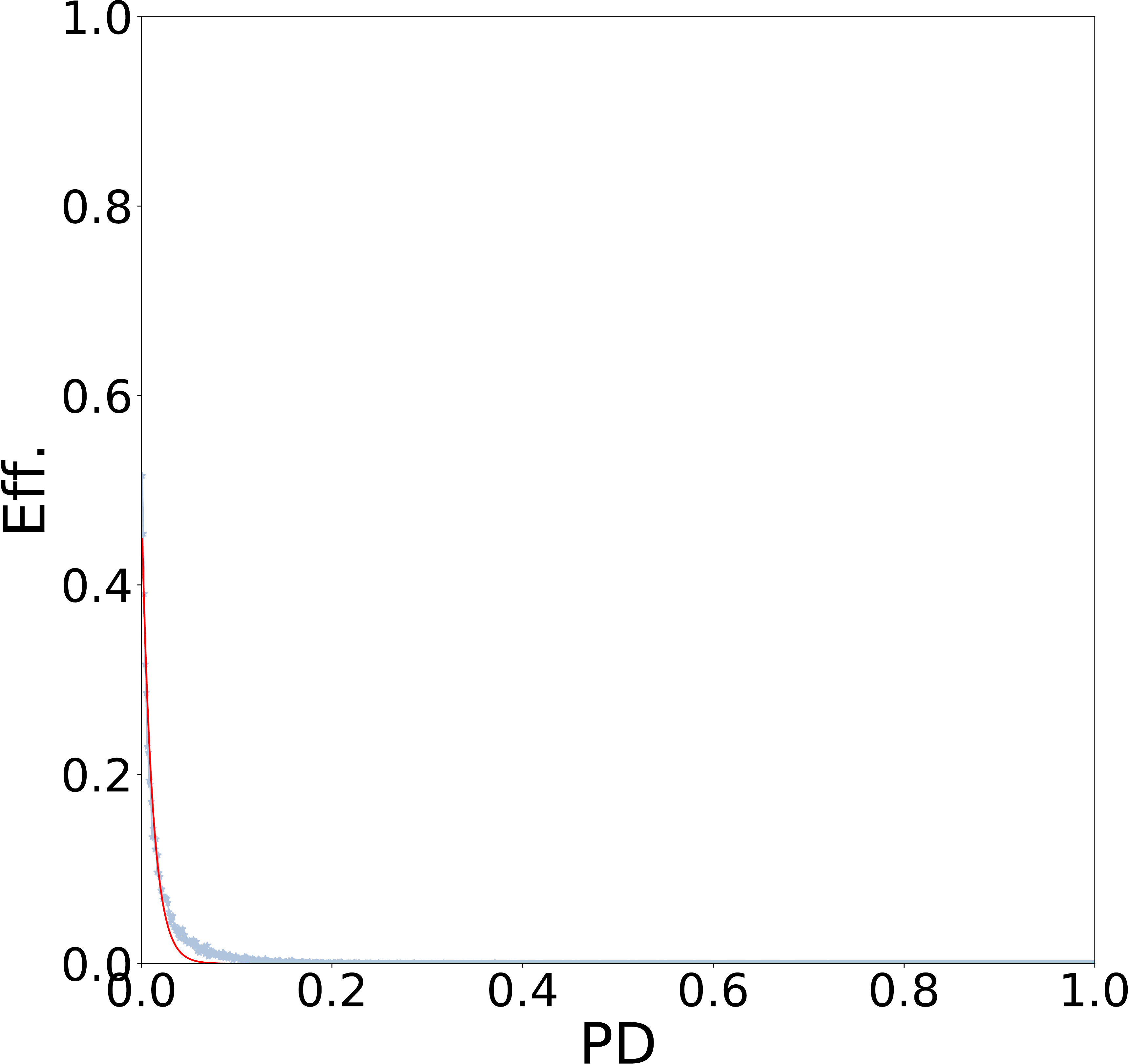}
\end{minipage}%
}%
\subfigure[f-MNIST on PD]{
\begin{minipage}[t]{0.31\linewidth}
\centering
\includegraphics[width=2.9cm,page=2]{figure_net.pdf}
\end{minipage}%
}
\subfigure[Cifar-10 on PD]{
\begin{minipage}[t]{0.31\linewidth}
\centering
\includegraphics[width=2.9cm,page=3]{figure_net.pdf}
\end{minipage}%
}
\\
\subfigure[Cifar-100 on PD]{
\begin{minipage}[t]{0.31\linewidth}
\centering
\includegraphics[width=2.9cm,page=4]{figure_net.pdf}
\end{minipage}%
}%
\subfigure[MNIST on PE]{
\begin{minipage}[t]{0.31\linewidth}
\centering
\includegraphics[width=2.9cm,page=5]{figure_net.pdf}
\end{minipage}%
}
\subfigure[f-MNIST on PE]{
\begin{minipage}[t]{0.31\linewidth}
\centering
\includegraphics[width=2.9cm,page=6]{figure_net.pdf}
\end{minipage}%
}
\\
\subfigure[Cifar-10 on PE]{
\begin{minipage}[t]{0.31\linewidth}
\centering
\includegraphics[width=2.9cm,page=7]{figure_net.pdf}
\end{minipage}%
}%
\subfigure[Cifar-100 on PE]{
\begin{minipage}[t]{0.31\linewidth}
\centering
\includegraphics[width=2.9cm,page=8]{figure_net.pdf}
\end{minipage}%
}
\subfigure[MNIST on PV]{
\begin{minipage}[t]{0.31\linewidth}
\centering
\includegraphics[width=2.9cm,page=9]{figure_net.pdf}
\end{minipage}%
}
\subfigure[f-MNIST on PV]{
\begin{minipage}[t]{0.31\linewidth}
\centering
\includegraphics[width=2.9cm,page=10]{figure_net.pdf}
\end{minipage}%
}
\subfigure[Cifar-10 on PV]{
\begin{minipage}[t]{0.31\linewidth}
\centering
\includegraphics[width=2.9cm,page=11]{figure_net.pdf}
\end{minipage}%
}
\subfigure[Cifar-100 on PV]{
\begin{minipage}[t]{0.31\linewidth}
\centering
\includegraphics[width=2.9cm,page=12]{figure_net.pdf}
\end{minipage}%
}
\caption{ The Results of Net-1}
\label{fig:resutlsNet1}
\end{figure}

\begin{table*}[htbp]
  \caption{Summary on Fitted Functions}
  \label{tab:fittedfunctions}
  \centering

  \begin{tabular}{l|c@{\ \ }c@{\ \ }c@{\ \ }|c@{\ \ }c@{\ \ }c}
  \toprule
  & \multicolumn{3}{ c |}{ Net-2} & \multicolumn{3}{ c }{ Net-1} \\
  \midrule
                & PD    &  PE   & PV    & PD    & PE  & PV  \\
  \toprule
        MNIST   & $f(x)=0.27 e^{-3.87x}$  &  $f(x)=0.27 e^{-3.81x}$ & $f(x)=0.27 e^{-3.86x}$  & $f(x)=0.49 e^{-90.37x}$  & $f(x)=0.43 e^{-74.28x}$ & $f(x)=0.47 e^{-84.28x}$    \\
 f-MNIST   & $f(x)=0.39 e^{-3.18x}$  &  $f(x)=0.38 e^{-3.11x}$ & $f(x)=0.39 e^{-3.16x}$  & $f(x)=0.47 e^{-13.27x}$  & $f(x)=0.42 e^{-10.80x}$ & $f(x)=0.45 e^{-12.31x}$    \\
    Cifar-10    & $f(x)=0.14 e^{-3.87x}$  &  $f(x)=0.14 e^{-4.00x}$ & $f(x)=0.14 e^{-3.95x}$  & $f(x)=0.42 e^{-20.76x}$  & $f(x)=0.22 e^{-8.88x}$ & $f(x)=0.27 e^{-11.46x}$    \\
    Cifar-100   & $f(x)=0.23 e^{-1.56x}$  &  $f(x)=0.24 e^{-1.64x}$ & $f(x)=0.24 e^{-1.61x}$  & $f(x)=0.43 e^{-8.52x}$  & $f(x)=0.31 e^{-5.42x}$ & $f(x)=0.36 e^{-6.53x}$    \\

  \bottomrule
  \end{tabular}
\end{table*}%

\begin{table}[tb]
  \caption{The Correlation Coefficient between Functions and Data Distribution}
  \label{tab:correlation}
  \centering
  \begin{tabular}{l|ccc|ccc}
  \toprule
  & \multicolumn{3}{ c |}{ Net-2} & \multicolumn{3}{ c }{ Net-1} \\
  \midrule
                & PD    &  PE   & PV    & PD    & PE  & PV  \\
  \toprule
        MNIST   & 0.99	&	0.99	&	0.99	&	0.97	&	0.97	&	0.97    \\
 f-MNIST        & 0.98	&	0.98	&	0.98	&	0.98	&	0.98	&	0.98    \\
    Cifar-10    & 0.98	&	0.97	&	0.98	&	0.97	&	0.97	&	0.97    \\
    Cifar-100   & 0.97	&	0.97	&	0.98	&	0.97	&	0.97	&	0.96    \\

  \bottomrule
  \end{tabular}
\end{table}%
To answer Q1, we trained two DNN models on four image datasets, and for given test examples
collected their noise sensitivity by three distance metrics, respectively. Moreover, for each given examples, we use attack models to generate 100 adversarial examples, and these adversarial examples are evaluated by Eff. metric. We use the collected data to generate the tuple set, and each tuple includes the distance value and the Eff. value for one example.

We visualize each tuple set in a x-y coordinate figure. In our experiment, there are four image datasets,three distance metrics and two DNN models, therefore, we can obtain 24 ($4\times 3 \times 2$) figures showed in Fig. \ref{fig:resutlsNet1} and Fig. \ref{fig:resutlsNet2}. Fig. \ref{fig:resutlsNet1} shows the Net-1 results on four image datasets, three distance metrics, respectively.  Fig. \ref{fig:resutlsNet1} shows the Net-2 results on four image datasets, three distance metrics, respectively.

Let us take Fig. \ref{fig:resutlsNet1} (a) for discussion first. Fig. \ref{fig:resutlsNet1} (a) shows the tuple distribution of MNIST on PD metric. The value of Eff. decreases as the value of PD increases. The distribution of Fig. \ref{fig:resutlsNet1} (a) conforms to the exponential distribution. We use least squares and variable substitution to fit a curve function with more than 95\% correlation coefficient between the function and data. We further observe other  distribution of each tuple set, which also conforms to the exponential distribution, and we fit a function with more than 95\% correlation coefficient for each dataset on each metric. all fitted functions  and the correlation coefficient are  summarized in Table \ref{tab:fittedfunctions} and Table \ref{tab:correlation}. We check all figures and obtain the relationship between Eff. values and the distance values is nonlinear-inverse proportion.

Finally, we answer $Q$1 as follows:
\begin{enumerate}[\ \ \ \textit{A}1:]
\item The examples are sensitive to noise, and the noise sensitivy and the distance of probability vectors are exponential relationships.
\end{enumerate}

\subsubsection{Answer Q2}

\begin{table}[tb]
  \caption{The results between NSATP and Random in F-measure}
  \label{tab:f-measure}
  \centering
  \begin{tabular}{l|l|c@{\ \ }c@{\ \ }c@{\ \ }|c@{\ \ }c@{\ \ }c}
  \toprule
  \multicolumn{2}{l|}{}& \multicolumn{3}{ c |}{ Net-2} & \multicolumn{3}{ c }{ Net-1} \\
  \midrule
  \multicolumn{2}{l|}{}& PD    &  PE   & PV    & PD    & PE  & PV  \\
  \toprule
       \multirow{3}*{MNIST}   &NSATP& 9.99	&	10.51	&	10.03	&	7.13	&	5.47	&	5.39    \\
                              &Radom & 908.52	&	918.60	&	900.90	&	144.82	&	136.68	&	135.00    \\
                              &Inc. & 98.90\%	&	98.86\%	&	98.89\%	&	95.07\%	&	96.00\%	&	96.01\%    \\
                              \midrule
 \multirow{3}*{f-MNIST}  &NSATP & 1.83	&	2.18	&	1.94	&	4.91	&	4.92	&	4.87    \\
                               &Radom & 488.45	&	490.23	&	503.65	&	92.48	&	93.27	&	97.49    \\
                               &Inc. & 99.63\%	&	99.56\%	&	99.61\%	&	94.69\%	&	94.72\%	&	95.01\%    \\
                               \midrule
  \multirow{3}*{  Cifar-10}   &NSATP & 1.94	&	6.89	&	4.26	&	20.43	&	14.98	&	15.01    \\
                                &Radom & 609.31	&	606.19	&	626.15	&	300.14	&	323.44	&	308.96    \\
                                &Inc. & 99.68\%	&	98.86\%	&	99.32\%	&	93.19\%	&	95.37\%	&	95.14\%    \\
                                \midrule
  \multirow{3}*{  Cifar-100}  &NSATP & 1.54	&	2.04	&	1.58	&	8.30	&	5.52	&	5.30    \\
                                  &Radom & 287.74	&	287.07	&	296.62	&	40.85	&	42.32	&	38.47    \\
                                  &Inc. & 99.46\%	&	99.29\%	&	99.47\%	&	79.67\%	&	86.96\%	&	86.23\%    \\

  \bottomrule
  \end{tabular}
\end{table}%
The effectiveness of NSATP and Random in revealing adversarial attack on one DNN are evaluated using the F-measure metrics. The experiment results of NSATP and Random in F-Measure on four image datasets and two models are shown in Table \ref{tab:f-measure}.
To answer Q2, we randomly select 1000 examples to simulated random testing\footnote{The Eff. values of all examples are more than zero.}, and select top 1000 examples of ranked example list. We collect the F-measure for these examples on two models, four image datasets, and three distance metrics.

Table \ref{tab:f-measure}  lists out the effectiveness of NSATP and Random in F-Measure, for each image dataset, each distance metric, and each model.
 Let us take the first cell to illustrate to the contents. It shows that the F-Measure for NSATP and Random on the Net-1 model, MNIST and PD metric are 2.5 and 5, respectively. It means that on average (recalling that ten individual tests are conducted for each testing to avoid sample bias), 2.5 and 5 payloads needed to be evaluated before an effective adversarial example is generated, by NSATP and Random, respectively. We further use the Inc. ($\frac{R-A}{R}\times 100\%$) metric to calculate the improvements from Random to ART4SQLi, and report Inc.$=50\%$. We check all Inc. values of \ref{tab:f-measure} and obtain they are range from 30\% to 60\%, which shows NSATP is more effective than Random.

Finally, we answer $Q$2 as follows:
\begin{enumerate}[\ \ \ \textit{A}2:]
\item Our method can effectively analyze the noise sensitivity of examples, and the fitted curves based on the results of three distance metrics are all exponential function on each image dataset and each model.
\end{enumerate}

\subsubsection{Answer Q3}

\begin{table}[htbp]
  \caption{Map the curve function to the linear function}
  \label{tab:map}
  \centering
  \begin{tabular}{l|l|c@{\ \ }c@{\ \ }c@{\ \ }|c@{\ \ }c@{\ \ }c}
  \toprule
  \multicolumn{2}{l|}{} & \multicolumn{3}{ c |}{ Net-1} & \multicolumn{3}{ c }{ LeNet-2} \\
  \midrule
   \multicolumn{2}{l|}{}             & PD    &  PE   & PV    & PD    & PE  & PV  \\
  \toprule
   \multirow{2}*{MNIST}                    &Abs.         & $3.87$  &  $3.81$ & $3.86$  & $90.37$ & $74.28$ & $84.28$    \\
                       &mark       & Best  & Worst  & Middle  & Best  &  Worst  &  Middle  \\
  \midrule
  \multirow{2}*{f-MNIST}                              &Abs.         & $3.18$  &  $3.11$ & $3.16$  & $13.27$ & $10.80$ & $12.31$    \\
                                &mark        & Best  & Worst  & Middle  & Best  &  Worst  &  Middle    \\
  \midrule
   \multirow{2}*{  Cifar-10}                             &Abs.         & $3.87$  &  $4.00$ & $3.95$  & $20.76$ & $8.88$ & $11.46$    \\
                                &mark        & Worst  &  Best & Middle  & Best  &  Worst &  Middle   \\
  \midrule
  \multirow{2}*{  Cifar-100}                          &Abs.         & $1.56$  &  $1.64$ & $1.61$  & $8.52$ & $5.42$ & $6.53$    \\
                            &mark        & Worst  &  Best & Middle  & Best  &  Worst & Middle    \\

  \bottomrule
  \end{tabular}
\end{table}%
To answer Q3, we use Eff-D metric to evaluate the effectiveness of the three distance metrics.

Let us take the first cell of Table \ref{tab:fittedfunctions} to illustrate the evaluation process. We define a function $z=\ln y$, and we use variable substitution to obtain the function $z=-4.64x+\ln 0.38$. We further compute the absolute derivative ($z'$) of z versus x, which is 4.64. We map each function of Table \ref{tab:fittedfunctions} to a linear function and compute the absolute derivative of the linear function, summarized in Table \ref{tab:map}. We compare the absolute derivatives on one model and one dataset among three distance metric, and mark the absolute derivative is maximum, middle, minimum as $Best$, $Middle$ and $Worst$, respectively.
We find that the probability difference has 6 ``Best", 0 ``Middle'' and 2 ``Worst'', the probability entropy has 2 ``Best", 0 ``Middle'' and 6 ``Worst'', and the probability variance has 0 ``Best", 8 ``Middle'' and 0 ``Worst'', which denotes that the probability difference is better the other two  metrics  to evaluate the noise sensitivity of examples in general.

Finally, we answer $Q$3 as follows:
\begin{enumerate}[\ \ \ \textit{A}3:]
\item The probability difference metric is better than the other two metrics on each image dataset and each model.
\end{enumerate}

\subsection{Threats to validity}
In this section, we discuss the threats to validity of our experiment.

In our experiment, we train two DNN models on four image datasets, and one perturbation modifies four ($2 \times 2$) pixels. Different perturbation may lead to different observations. The  perturbation strategy just modifies four pixels, but other perturbation strategies  are likely to modify more than four pixels. The examples with too many noise may be easily detected by adversarial defense mechanism. Moreover, we randomly choose the pixels of given examples and modify them. To enhance the observations and results, we repeat the experiment for 3 times, and their conclusions are consistent.

Different models may have different accuracy rates with different training strategies, and different trained models may affect the effectiveness of our method. In our experiment, we construct two different models and different training strategies. The results based on these models and training strategies are consistent.

\section{Related Work}
\label{sec:relatedwork}
Machine learning is increasingly being used for potential safety critical decisions, such as malware classfication, autopilot system, face recognition, financial transaction \& supervision, and so on.  It is  essential to prevent the potential errors from being exposed. Among all the deep learning techniques, deep neural networks (DNNs) are one of the most popular techniques and have shown excellent results in  speech  recognition \cite{hinton2012deep,sainath2013deep}, image recognition \cite{schmidhuber2015deep}, natural language processing \cite{collobert2008unified}.

Machine learning models are often vulnerable to adversarial manipulation of their input intended to cause incorrect classification \cite{dalvi2004adversarial}.
DNNs are actually an effective data representation of a given set of samples. A trained DNN models  are highly vulnerable to attacks based on adversarial examples with small modifications.
Szegedy et al. \cite{szegedy2013intriguing} first revealed the sensitivity to well-tuned artificial perturbation, such perturbation image can fool DNN models, they further observed that  using adversarial examples to train DNN can improve the robustness of DNN against adversarial examples.  Goodfellow et al. \cite{goodfellow2014generative} proposed a framework to estimate generative models via an adversarial process, which is called generative adversarial networks (GANs).  Kurakin et al. \cite{kurakin2016adversarial} proposed basic iterative method to generate adversarial examples, which add little perturbation in each step. Papernot et al. \cite{papernot2016limitations}  formalized the space of adversaries against deep neural networks and
utilize Jacobian matrix to build adversarial saliency map to  enable an efficient exploration of the adversarial-samples search space.
Su et al \cite{su2017one}  just modified one pixel of examples and observed the adversarial examples can successful fool DNNs.
Carlini and Wagner \cite{carlini2017towards} introduced three adversarial attacks in the wake of defensive distillation against
the adversarial perturbations.
Moosavi-Dezfooli et al. \cite{moosavi2016deepfool} proposed to compute a minimal norm adversarial perturbation for a given image in an iterative manner.
Akhtar and Mian \cite{akhtar2018threat} have investigated other adversarial attacks. Our method is to evaluate the noise sensitivity of examples, which can work on these attacks.

\section{Conclusion}
\label{sec:conclusion}

Deep neural networks (DNNs) have been widely use in natural language processing, computer vision and image recognition. Many researchers have proposed many efficient DNNs for those applications. But, related works has show the DNNs can be fooled by  artificial examples with some perturbations and many adversarial attack models have be proposed. However, most of them pay little attention to the noise sensitivity of examples.

In this paper, we have studied the correlation between the noise sensitivity  and the probability labels of examples. We proposed a noise-sensitivity-analysis-based test prioritization technique for deep neural networks, which evaluates the the noise sensitivity of examples based on the probability labels of their examples. We have conducted a controlled experiment on two  deep neural networks over four image datasets with three distance metrics. The experimental result has confirmed examples are sensitive to noise. Empirical results have shown that NSATP can effective evaluate the noise sensitivity of examples and rank the examples by their noise sensitivity, and probability-difference distance is more effective than probability-entropy distance and probability-variance distance.

Future work includes a thorough study on other example datasets and deep learning models. In order to enhance the effectiveness of our method, another future work is to validate it on other adversarial attack models as well.

\section{Acknowledgements}
This work was supported by grant from the XXX, grant from the XXX, and grant from the XXX.

\bibliographystyle{IEEEtran}

\vfill
\clearpage
\end{document}